\documentclass[10pt,twocolumn,letterpaper]{article}

\usepackage{cvpr}
\usepackage{times}
\usepackage{epsfig}
\usepackage{graphicx}
\usepackage{amsmath}
\usepackage{pifont}	
\usepackage{amssymb}
\usepackage{booktabs}
\usepackage{algorithm}
\usepackage{algorithmicx}
\usepackage{algpseudocode}

\usepackage{multirow}
\usepackage{array}

\usepackage[pagebackref=true,breaklinks=true,letterpaper=true,colorlinks,bookmarks=false]{hyperref}

\cvprfinalcopy 

\def\cvprPaperID{874} 

\ifcvprfinal\fi
\begin{document}

\title{Deep Group-shuffling Random Walk for Person Re-identification}

\author{
Yantao Shen$^{1}$ \quad 
Hongsheng Li$^{1}$\thanks{H. Li and X. Wang are the co-corresponding authors.} \quad 
Tong Xiao$^{1}$ \quad 
Shuai Yi$^{2}$ \quad 
Dapeng Chen$^{1}$ \quad
Xiaogang Wang$^{1}$\footnotemark[1] \\
$^{1}$ CUHK-SenseTime Joint Lab, The Chinese University of Hong Kong\\
$^{2}$ SenseTime Research\\
$^{1}${\tt\small \{ytshen, xiaotong, hsli, dpchen, xgwang\}@ee.cuhk.edu.hk  }\\
$^{2}${\tt\small yishuai@sensetime.com} \\
}


\maketitle

\begin{abstract}
Person re-identification aims at finding a person of interest in an image gallery by comparing the probe image of this person with all the gallery images. It is generally treated as a retrieval problem, where the affinities between the probe image and gallery images (P2G affinities) are used to rank the retrieved gallery images. However, most existing methods only consider P2G affinities but ignore the affinities between all the gallery images (G2G affinity). Some frameworks incorporated G2G affinities into the testing process, which is not end-to-end trainable for deep neural networks. In this paper, we propose a novel group-shuffling random walk network for fully utilizing the affinity information between gallery images in both the training and testing processes. The proposed approach aims at end-to-end refining the P2G affinities based on G2G affinity information with a simple yet effective matrix operation, which can be integrated into deep neural networks. Feature grouping and group shuffle are also proposed to apply rich supervisions for learning better person features. The proposed approach outperforms state-of-the-art methods on the Market-1501, CUHK03, and DukeMTMC datasets  by large margins, which demonstrate the effectiveness of our approach.
\end{abstract}

\section{Introduction}
\label{sec:intro}

Person re-identification (Re-ID) is a challenging problem. Given one probe image of a person of interest, the task requires to identify  images of the same person from a large gallery image database. It is an important and active research field and has vital roles in video surveillance systems. In recent years, deep learning methods achieved huge success in various computer vision tasks. There have been attempts on solving the person re-ID task with deep learning methods, which focus on learning discriminative feature representations of person images. The re-identification problem is solved as an image retrieval task, where the gallery images are ranked according to their affinities (\eg, Euclidean distances between image features) to the probe image. Such probe-to-gallery affinities are called \emph{P2G affinities} in this paper.

\begin{figure}[t]
\centering
\small
\begin{tabular}{c@{\hspace{0mm}}c}
   &\includegraphics[scale=0.4]{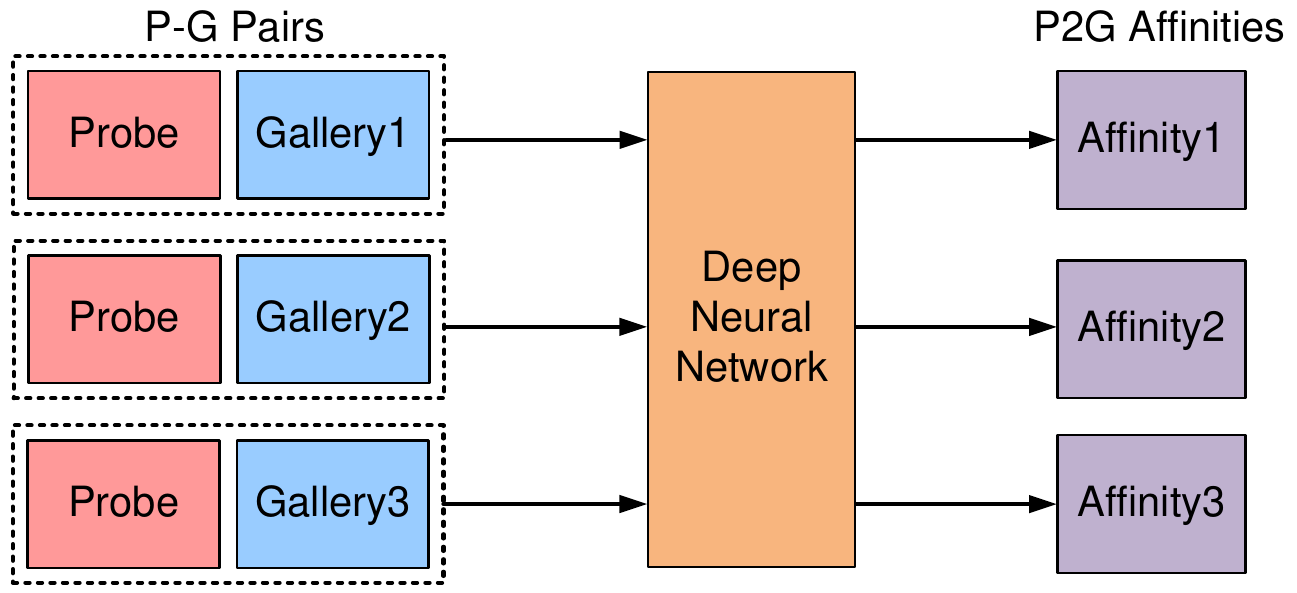}\\
   & (a) Conventional approaches for estimating P2G affinities \\
   &\includegraphics[scale=0.4]{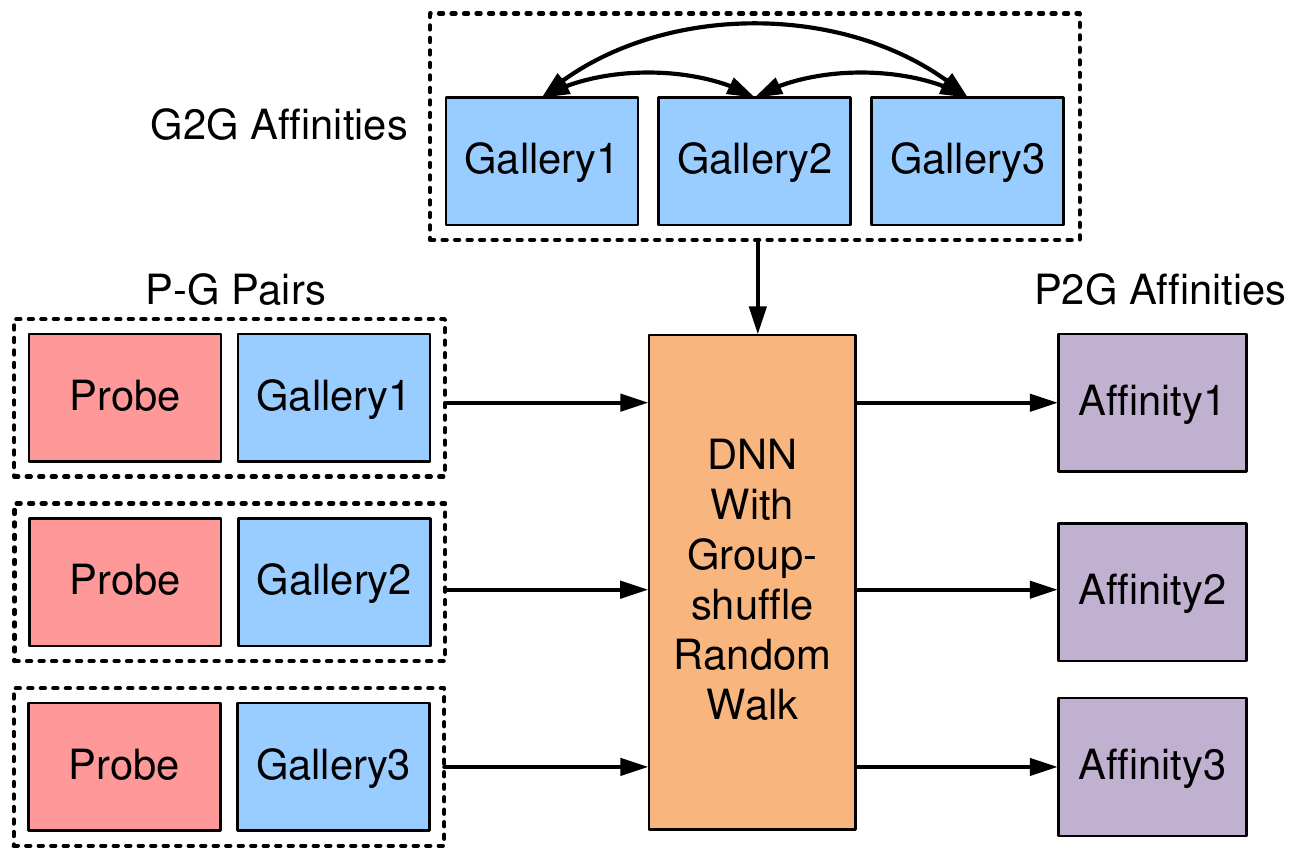}\\
   &(b) Proposed approach
\end{tabular}
   \caption{(a) Most conventional approaches only utilize information between pairs of probe and gallery images for P2G affinity estimation. (b) Proposed approach with end-to-end group-shuffling random walk integrates G2G affinities for P2G affinity estimation.}
\label{fig:path}
\vspace{-12pt}
\end{figure}

However, relying only on P2G affinities to rank the gallery images is not robust enough. For instance, if the probe image shows a person's frontal view. When comparing with the same person's back-view image, it is unlikely to obtain a high affinity score with the probe due to the large viewing angle difference. However, if there exists a side-view image of the person, which has high affinities with both the same person's frontal-view and back-view images. The frontal-view and back-view images could then be matched with high confidence. This indicates that the affinities between gallery images (named \emph{G2G affinities}) are valuable for determining the final P2G affinities between the probe and gallery images.

Incorporating G2G affinities to improve the initial ranking of gallery images is considered as a re-ranking problem. There were some previous attempts on re-ranking with affinities between top-ranked gallery images~\cite{Zhong_2017_CVPR, garcia2015person, leng2015person, ye2015ranking, ye2016person}. Most of these re-ranking approaches utilized the $k$-nearest neighbors of the gallery image. They assumed that if the probe image is contained in the nearest neighbor set of a gallery image, their affinity should be large. There were also manifold ranking methods \cite{loy2013person, bai2017scalable} for re-ranking gallery images. The G2G affinities between gallery images are incorporated to refine the initial P2G affinities based on the random walk algorithm. However, all above mentioned methods conduct the re-ranking as a separate post-processing stage. The affinities between gallery images are not taken into account for better learning features in the training phase.


To address the problem, we propose a novel group-shuffling random walk (GSRW) layer for deep neural networks, which integrates the random walk operation in both training and testing process for generating accurate probe-to-gallery affinities and discriminative person features. Given a probe and a group of gallery images, the neural network first generates the initial P2G and G2G affinities between them. The GSRW layer takes the affinities as inputs and propagates information among images via the random walk operation to generate the refined P2G affinities. For better training individual feature dimensions, the feature dimensions are divided into several groups to generate multiple groups of initial P2G and G2G affinities. By applying the random walk algorithm with the pairwise combinations of the grouped P2G and G2G affinities, the person feature learning is better regularized. Extensive experiments on three public datasets demonstrate the effectiveness of our proposed approach and the individual components.

The contribution of this paper is threefold. (1) We propose a novel group-shuffling random walk layer that integrates the P2G and G2G affinities to obtain more accurate probe-to-gallery affinities. Unlike existing methods that treat re-ranking as a separate post-process stage, the proposed GSRW layer can be end-to-end trained within deep neural networks and results in more discriminative feature representations. (2) We propose to divide feature dimensions into several groups and apply supervision signals separately to each group. This simple strategy could force each feature dimension to contribute to capturing discriminative information for affinity estimation. (3) Based on the group feature sub-vectors, we propose a group-shuffling operation that combines multiple pairs of initial P2G and G2G affinities for training. This operation implicitly applies rich supervision signals on both P2G and G2G affinities and better regularizes the feature learning process.

\begin{figure*}[t]
   \begin{center}
      \includegraphics[scale=0.6]{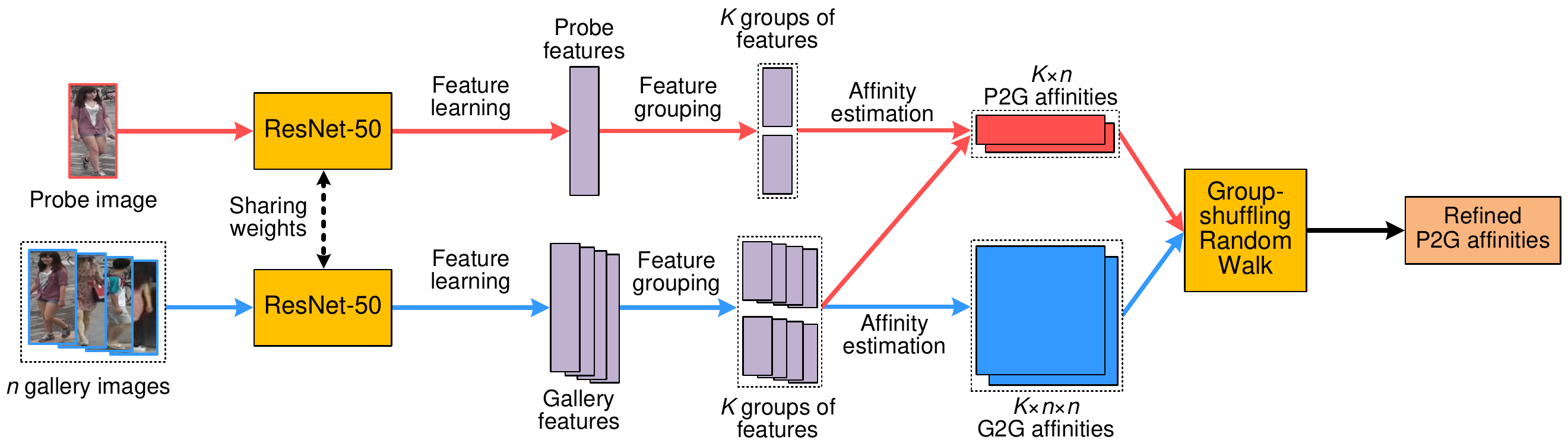}
   \end{center}{}
   \vspace{-10pt}
   \caption{Illustration of the proposed approach with group-shuffling random walk operation. Given a probe and a set of gallery images, their initial P2G and G2G affnities are estimated by a pairwise affinity CNN. With our proposed feature grouping and group-shuffling random walk, the P2G affinities are refined as the final results.}
   \label{fig:main}
   \vspace{-7pt}
\end{figure*}

\section{Related Work} 
\label{sec:related_work}

\textbf{Deep learning based person re-identification.} In recent years, person re-identification gains increasing attention from both industry and academia~\cite{loy2013person, ahmed2015improved, li2014deepreid,bkak2016person,Zhong_2017_CVPR,shen2017learning,shen2018end}. It is a challenging computer vision task because of the drastic variations of human poses, various camera views, and occlusions. With the emergence of deep learning techniques, state-of-the-art person re-identification methods adopted Convolutional Neural Networks (CNN) for learning person features. Li \etal~\cite{li2014deepreid} designed a filter pairing neural network to jointly handle misalignment and geometric transformations. Ahmed \etal~\cite{ahmed2015improved} proposed a Cross-Input Difference CNN to capture local relationships between the two input images based on mid-level features from each input image. Ding \etal~\cite{ding2015deep} exploited triplet samples for training CNNs to minimize the feature distance between positive samples and maximize the distance between negative samples. Xiao \etal~\cite{xiao2016learning} proposed a Domain Guided Dropout technique to mitigate the domain gaps between different person Re-ID datasets. Chen \etal~\cite{Chen_2017_CVPR} proposed quadruplet loss to train a deep network, which aims to learn features with large inter-class variations and smalle intra-class variations. Zhao \etal~\cite{zhao2017spindle} and Su \etal~\cite{Su_2017_ICCV} integrated human pose information for tackling the pose variation problem and improving feature learning capability. Besides deep learning based person Re-ID methods, a large number of metric learning based approaches~\cite{Bak_2017_CVPR,Liu_2017_ICCV,Yu_2017_ICCV,Zhang_2016_CVPR} were also proposed to learn better distance metrics to measure similarities between person images.

\textbf{Re-ranking for person re-identification.} There were some preliminary attempts on incorporating affinities between gallery images into the ranking process~\cite{liu2013pop, wang2016human, ye2015ranking, ye2016person, Zhong_2017_CVPR, bai2017scalable, loy2013person}. Some approaches require human interaction~\cite{liu2013pop, wang2016human}, which are not automatic and label-free. Ye \etal~\cite{ye2016person} utilized local and global features as additional probes. The initial ranking is improved by integrating new ranking of the local and global features. Zhong \etal~\cite{Zhong_2017_CVPR} exploited $k$-reciprocal neighbors in person Re-ID. Compared with $k$-nearest neighbors, the $k$-reciprocal neighbors of gallery image are more related to the probe image. Furthermore, to avoid calculating Jaccard Distance between $k$-reciprocal neighbors sets of probe and gallery, a feature distance equivalent to Jaccard Distance was proposed, which could be computed in parallel on GPUs. 

Directly propagating the G2G affinities to P2G affinities was also proposed to refine P2G affinities. Loy \etal~\cite{loy2013person} and Bai~\etal~\cite{bai2017scalable} adopted random walk operation to adjust P2G affinities.  Compared with~\cite{loy2013person}, ~\cite{bai2017scalable} exploited the training data and their labels to output the revised matching probabilities. However, all the above methods only conducted re-ranking as a separate post-processing stage during testing, which is not end-to-end trainable and cannot help to learn better features. In contrast, our proposed framework integrates the novel group-shuffling random walk operation into deep neural networks, which benefits feature learning and significantly improves test accuracy.

\textbf{Random walk algorithms.} Random walk is a well-known graphical model~\cite{aldous2002reversible}. It has extensive applicatons in webpage ranking \cite{page1999pagerank} and image segmentation \cite{bertasius2016convolutional}. Bertasius \etal~\cite{bertasius2016convolutional} incorporated random walk in deep neural networks for image segmentation. An affinity learning branch is designed to regularize the pixel prediction results based on inter-pixel affinities. This method considered pixel-to-pixel relations within a single image, while our proposed method focuses on using inter-image relations for improving image affinity ranking.
As discussed above, the random walk algorithm~\cite{loy2013person}, ~\cite{bai2017scalable} was also used as a post-processing step for person re-ID but was not end-to-end trained with deep neural networks.

\section{Approach}
\label{sec:approach}

Given a probe person image and multiple gallery images, the goal of person re-ID is to estimate accurate affinity scores between the probe image and gallery images (P2G affinities), which represent the probabilities that each probe-gallery image pair belong to the same person. The gallery images could then be ranked according to the P2G affinities as the final results. As shown in Figure \ref{fig:main}, our proposed approach aims at exploiting the similarities between gallery images (G2G affinities) to improve the accuracy of the initial P2G affinities. This is achieved by integrating a novel group-shuffling random walk layer into an end-to-end trainable deep neural network for learning discriminative features and accurately estimating P2G affinities.

The basis of the random walk algorithm will be reviewed in Section \ref{ssec:random_walk_background}. We then present the end-to-end random walk operation in deep neural networks for person re-ID in Section \ref{ssec:end_to_end_random_walk}, and discuss the rich supervisions brought by integrating the random walk algorithm in Section \ref{ssec:rich_supervisions}. The feature grouping and group shuffle for further boosting the feature learning is introduced in Section \ref{ssec:group_shuffling_random_walk}.

\subsection{Random walk algorithm} 
\label{ssec:random_walk_background}

The random walk algorithm~\cite{aldous2002reversible} is well-known for being the foundation of the PageRank algorithm~\cite{page1999pagerank} on webpage ranking. Let $G=(V, E)$ denote an undirected graph where $V$ denotes the vertices and $E$ denotes the edges. The random walk operation on the graph can be modeled with an $n \times n$ square matrix $W$, where $n$ is the number of vertices. $W(i,j) \in [0, 1]$ can be viewed as the similarity probability between the $i$-th and $j$-th nodes with constraints $\sum_j W(i,j) = 1$ for all $j$. In the context of person Re-ID, $W(i,j)$ can be considered as the normalized affinity scores between the $i$-th and $j$-th person gallery images and each gallery image is a node on the graph.

Given the probe image and $n$ gallery images. Let $y^{(t)}$ be an $n \times 1$ vector denoting the P2G affinity scores between the probe image and all gallery images at random walk iteration $t$. With the matrix $W$ storing normalized pairwise affinities between pairs of gallery images, the random walk operation can be characterized as $y^{(t+1)} = W y^{(t)}$, where $y^{(t+1)}$ denotes the refined P2G affinities at iteration $t+1$. Such operation diffuses the information of G2G affinities to P2G affinities to refine the P2G affinities. The iteration can be conducted recursively until the refined P2G affinities converge.

\subsection{Random walk in deep neural networks} 
\label{ssec:end_to_end_random_walk}

In this section, we introduce the integration of the random walk operation into deep Convolution Neural Networks (CNN) for learning more discriminative features and for estimating accurate P2G affinities with the assistance of G2G affinities.

Given a probe image and a set of $n$ gallery images, the pairwise affinity scores between images could be estimated by a state-of-the-art person re-ID Siamese-CNN, which takes a pair of images as inputs and estimates the probability that the two images belong to the same person. We call the affinity scores between the probe and gallery images by the CNN the \emph{initial P2G affinities} $y^{(0)} \in \mathbb{R}^n$. We introduce a random walk layer for deep neural networks, which takes the initial P2G affinities and G2G affinities as inputs and outputs the refined P2G affinities. Let $S \in \mathbb{R}^{n \times n}$ denote the matrix that stores original G2G affinity scores between the $n$ gallery images. To fulfill the normalization constraints that $\sum_j W(i,j) = 1$ for all $j$, we normalize each row of the original affinity matrix $S$ with a softmax function, \ie,
\begin{align}  
	W(i,j) = \frac{\exp(S(i,j))}{\sum_{j \neq i} \exp({S(i,j))}}, \quad \text{for all $i = 1,\cdots, n$}.
	\label{eq:softmax}
\end{align}
Note that all the diagonal entries of $W$ are set to zero, \ie, $W(i,i) = 0$, for avoiding self-reinforcement during random walk iterations. Therefore, the diagonal entries are not involved in the softmax normalization in Eq. (\ref{eq:softmax}). The one iteration of random walk refinement on the initial P2G affinities can be formulated as
\begin{equation}
	y^{(1)} = Wy^{(0)},
	\label{eq:random_walk_one_step}
\end{equation}
where $y^{(1)}$ is the refined P2G affinities based on the initial P2G affinities $y^{(0)}$ and normalized G2G affinities $W$. Intuitively, if gallery images $i$ and $j$ are similar, their P2G affinities to the probe image should also be similar. For the $i$th image's refined affinity score $y^{(1)}(i)$, it is calculated as
\begin{align}
	y^{(1)}(i) = W(i,1) \cdot y^{(0)}(1) + \cdots + W(i,n) \cdot y^{(0)}(n).
\end{align}
From the equation, we can see that if gallery images $i$ and $j$ are more similar ($W(i,j)$ is large), the P2G affinity of image $j$, $y^{(0)}(j)$, has a higher weight $W(i,j)$ to be propagated to the P2G affinity score $y^{(1)}(i)$. 

In practice, we would like the refined P2G affinities not to deviate too far away from the initial P2G affinity estimations. We therefore weightedly combine the random walk refinements $y^{(1)}$ with the initial P2G affinity $y^{(0)}$ as
\begin{align}
	y^{(1)} = \lambda W y^{(0)} + (1-\lambda) y^{(0)},
\end{align}
where $\lambda \in [0,1]$ is the weighting parameter that balances the two terms. The random walk operation is generally conducted multiple times until convergence,
\begin{align}
y^{(t+1)} = \lambda W y^{(t)} + (1 - \lambda) y^{(0)},
\label{eq:weight_raondom_walk}
\end{align}
where $t$ represent the $t$th iteration. Expanding Eq. (\ref{eq:weight_raondom_walk}) leads to
\begin{equation}\label{eq:iter_rw_ex}
   y^{(t+1)} =(\lambda W)^{t+1}y^{(0)} + (1 - \lambda)\sum_{i=0}^{t}(\lambda W)^{i}y^{(0)}.
\end{equation}
As $t\rightarrow \infty$, since $\lambda \in [0,1]$,
\begin{align}
	\underset{t\rightarrow\infty}{\lim}(\lambda W)^{t+1}y = 0
\end{align}
For $\sum_{i=0}^{t}(\alpha W)^{i}$, the matrix series can be expanded as
\begin{equation}
   \underset{t\rightarrow \infty}{\lim} \sum_{i=0}^{t}(\lambda W)^{i}= (I - \lambda W)^{-1}.
\end{equation}
Eq. (\ref{eq:weight_raondom_walk}) could then be formulated as
\begin{align}
	y^{(\infty)} = (1 - \lambda)(I - \lambda W)^{-1}y^{(0)},
	\label{eq:random_walk}
\end{align}
where $I$ is the identity matrix. The calculation of Eq. (\ref{eq:random_walk}) can be modeled as a neural layer. By combining it with deep neural networks, it could be trained with the networks via back-propagation in an end-to-end manner. The supervisions could be directly applied to the output $y^{(\infty)}$, where each of its element represents the similarity probability of the probe image matching one of the gallery images.

\subsection{Rich supervisions from random walk}
\label{ssec:rich_supervisions}

Integrating the random walk layer into the deep neural networks not only helps propagate G2G affinities between gallery images to refine the P2G affinities, more importantly, it also provides rich supervisions for training the visual features of the input images.

Let $L(i)$ denote the prediction loss of predicting the affinity between the probe features $p$ and the $i$-th gallery image's features $G(i)$. Given the single affinity prediction error $L(i)$, conventional Siamese-CNNs only back-propagate the prediction errors to the two related images. In contrast, by introducing the random walk operation into the deep neural networks, the error $L(i)$ would be back-propagated to all P2G and G2G affinities, $y^{(0)}$ and $W$, providing rich supervisions for learning discriminative visual features.

To show it, we analyze the gradients of the error $L(i)$ w.r.t. all P2G and G2G affinities. 
The affinity score $y^{(0)}(i)$ of $p$ and $G(i)$ is computed as
\begin{equation}\label{eq:p2gscore}
	y^{(0)}(i) = h(p - G(i)), 
\end{equation}
where $h$ denotes a non-linear function for computing the affinity score (\eg, Euclidean distance). 
The gradients of $L(i)$ w.r.t. $y^{(0)}$ is calculated as
\begin{equation}\label{eq:rw_gradient_y_i}
   \frac{\partial L(i)}{\partial y^{(0)}} =
                                          \left[ 
                                          \begin{array}{ccc} 
                                             \frac{\partial L(i)}{\partial y^{(\infty)}(i)}\widehat{W}(i,1), \quad 
                                             \cdots, \quad 
                                             \frac{\partial L(i)}{\partial y^{(\infty)}(i)}\widehat{W}(i,n) \quad
                                          \end{array} 
                                          \right]^T,
\end{equation}
where $\widehat{W}$ represents $(1 - \lambda)(I - \lambda W)^{-1}$.

The gradients of $L(i)$ w.r.t all G2G affinities $W(i,j)$ can be calculated as 
\begin{equation}\label{eq:rw_gradient_W}
   \frac{\partial L(i)}{\partial W(i,j)} = -tr \left(\lambda \frac{\partial L(i)}{\partial y^{(\infty)}} y^{(0)T} \widehat{W} E^{ij} \widehat{W} \right),
\end{equation}
where $E^{ij}$ denotes an $n\times n$ matrix with 1 at entry $(i,j)$ and 0s elsewhere. 
Both Eqs. (\ref{eq:rw_gradient_y_i}) and (\ref{eq:rw_gradient_W}) demonstrate that the error $L(i)$ for a single probe-gallery image pair would back-propagate to all P2G affinities $y^{(0)}$ and all G2G affinities $W$, and further to visual features of the probe and all gallery images, $p$ and $G(1), \cdots, G(n)$.

In contrast, for conventional Siamese-CNNs which do not involve the random walk operation. The loss $L(i)$ would be back-propagated to only the P2G affinity $y^{(0)}(i)$ as
\begin{equation}\label{eq:no_rw_gradient_y_i}
   \frac{\partial L(i)}{\partial y^{(0)}} =
                                          \left[ 
                                          \begin{array}{ccccc} 
                                             0, \quad
                                             \hdots, \quad 
                                             \frac{\partial L(i)}{\partial y^{(0)}(i)}, \quad
                                             \hdots, \quad
                                             0
                                          \end{array} 
                                          \right]^T,
\end{equation}
and $\frac{\partial L(i)}{\partial W(i,j)} = 0$ for all $i,j = 1, \cdots, n$, providing much less supervision information for feature learning. 


\subsection{Group-shuffling random walk} 
\label{ssec:group_shuffling_random_walk}

As shown in the previous subsection, with the random walk layer, even the error of predicting a single probe-gallery pair's affinity is shown to provide rich supervisions to all images. However, the supervisions are in image-level and are applied to their whole visual feature vectors. During training, the training data might overfit the visual neurons in the neural network. Some neurons (feature dimensions) might be always inactive. For instance, for person images whose upper-body regions are more distinct than their lower-body regions. After training, the neurons for upper-body are well trained and those for lower-body might be always inactive because the upper-body features dominate the loss computation.
One possible solution is the dropout technique \cite{srivastava2014dropout}, which randomly drops the responses of a portion of neurons at each training iteration. Based on the property of the random walk layer, we propose a novel group-shuffling operation to solve this problem, which is shown to be complementary to the dropout technique.

We first divide all persons visual features (neurons) into $K$ different groups along the feature dimension. The visual features of the probe image $p$ are divided into feature sub-vectors $p_1, \cdots, p_K$. Similarly, $G_k(i)$ denotes the $k$th feature sub-vector of $G(i)$ and $W_k$ denotes the normalized affinity matrix of the $k$th feature group. Instead of directly predicting the affinity based on the whole feature vectors, $p$ and $G(i)$, as in Eq. (\ref{eq:p2gscore}), we now require to predict the affinity scores based on each feature group with much fewer number of features, \ie
\begin{align}
	y^{(0)}_k(i) = & h \left( p_k - G_k(i) \right) \,\, \text{ for } k = 1, \cdots, N.
\end{align}
In this way, the prediction tasks are more challenging and each feature dimension has a greater chance to contribute to the affinity prediction. We could apply the random walk operation to each feature group by substituting $y^{(0)}$ and $W$ in Eq. (\ref{eq:random_walk}) with $y^{(0)}_k$ and $G_k$. 

As shown in Eq. (\ref{eq:random_walk}), to make accurate predictions on the final P2G affinity scores $y^{(\infty)}_k$, it is important that both $y^{(0)}_k$ and $W_k$ are accurate. Otherwise the errors would be back-propagated to update them. Since $y^{(0)}_k$ and $W_k$ only represent affinity scores, $y^{(0)}_k$ and $W_k$ from different feature groups can be pairwisely combined. For instance, if $K=2$, we create 4 pairs of $\{y^{(0)}_1, W_1\}$, $\{y^{(0)}_1, W_2\}$, $\{y^{(0)}_2, W_1\}$, $\{y^{(0)}_2, W_2\}$ as inputs for the random walk layer to generate the refined P2G affinities $y_{11}^{(\infty)}$, $y_{12}^{(\infty)}$, $y_{21}^{(\infty)}$, $y_{22}^{(\infty)}$. The supervisions are independently applied to the refined P2G affinities of each combined pair. The group-shuffling operation is able to generate rich supervisions for training all feature groups. For instance, even if only the $2$nd feature group is not well trained, all $y_{12}^{(\infty)}$, $y_{21}^{(\infty)}$, $y_{22}^{(\infty)}$ would have large prediction errors to generate much supervisions for training the $2$nd feature group. The algorithm for group-shuffling random walk is illustrated in Algorithm \ref{alg:group_shuffling}.

\begin{algorithm}[h]
  \caption{Group-shuffling random walk}
  \begin{algorithmic}[1]
  
   \Require
   	  probe features $p$;
   	  features of $n$ gallery images $G(1), \cdots, G(n)$;
      group number $K$;
   \Ensure  
      Refined P2G affinities $y^{(\infty)}_{jk}$ for $j,k=1$,$\cdots,K$;
   
   \State Divide $p$ into $K$ groups, $p_1, \cdots, p_K$;
   \State Divide $G$ into $K$ groups, $G_1, \cdots, G_K$;
   \For{$k=1$ to $K$}
      \State Calculate P2G affinities $y^{(0)}_k$ with $p_k$ and $G_k$;
      \State Calculate G2G affinities $W_k$ with $G_k$;
   \EndFor
   \For{$j=1$ to $K$}
      \For{$k=1$ to $K$}
         \State $y_{jk}^{(\infty)} = (1 - \lambda)(I - \lambda W_j)^{-1}y^{(0)}_k$;
      \EndFor
   \EndFor

    \label{code:shuffle}
  \end{algorithmic}
  \label{alg:group_shuffling}
\end{algorithm}

\begin{figure}[t]
   \begin{center}
      \includegraphics[scale=0.4]{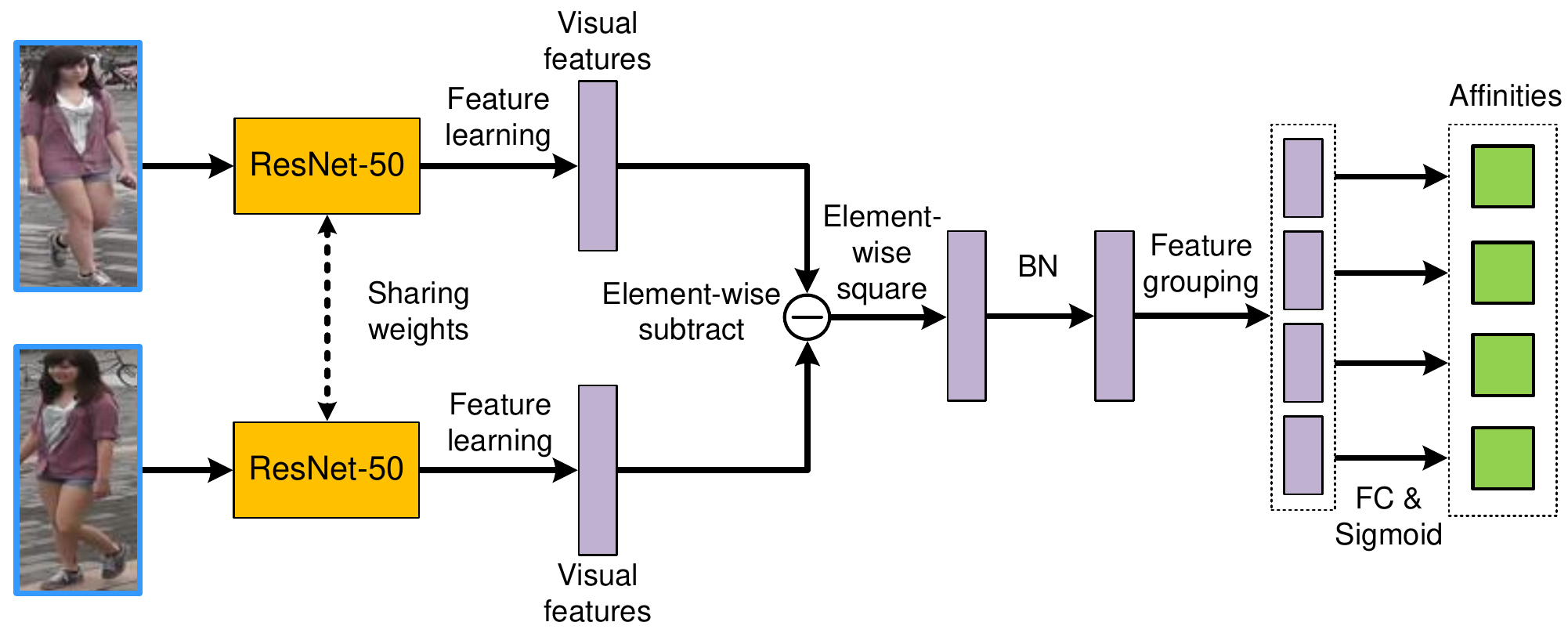}
   \end{center}{}
   \vspace{-10pt}
   \caption{Illustration of the pairwise affinity CNN. The resulting feature vector is divided into several groups, each of which is mapped to an affinity score.}
   \label{fig:pairwise}
   \vspace{-15pt}
\end{figure}

\subsection{Overall network structure} 
\label{ssec:overall_network_structure}

The overall deep neural network is illustrated in Figure \ref{fig:main}. It consists of a pairwise affinity CNN and the proposed group-shuffling random walk layer. 

The pairwise affinity CNN takes a pair of images as inputs and outputs $K$ affinity scores between the two images for $K$ feature groups. The network structure for the pairwise affinity CNN is shown in Figure \ref{fig:pairwise}. The Siamese part adopts the ResNet-50 \cite{he2016deep} structure until the global pooling layer. The two 2048-d feature vectors of the two images are then subtracted and processed by elementwise square and Batch Normalization \cite{ioffe2015batch}. The final feature vector is divided into $K$ sub-feature vectors, each of which is mapped to an affinity score by a fully-connected (FC) layer and a sigmoid function. Note that dividing the features in the last  layer is equivalent to dividing the output features from the average pooling layer. Given a probe image and $n$ gallery images, the pairwise affinity CNN estimates initial P2G affinities $y^{(0)}_k \in \mathbb{R}^n$ and G2G affinities $W_k \in \mathbb{R}^{n\times n}$ for $k=1,\cdots, K$. The group-shuffling random walk layer takes the initial P2G and G2G affinities as inputs and output $K^2$ groups of refined P2G affinities $y^{(\infty)}_{jk}$ for $j,k = 1,\cdots, K$. The supervisions are applied to the refined P2G affinities with cross-entropy loss functions.

%

\section{Experiments}
\label{sec:experiments}

To validate the effectiveness of our proposed approach on person Re-ID, we conduct experiments and ablation studies on three public datasets. 


\subsection{Datasets and metric} 
\label{sub:datasets_and_metric}

\textbf{Datasets.} 1) Market-1501~\cite{zheng2015scalable} consists of 12,936  images for training and 19,732 images for testing. There are 1,501 different persons in this dataset, which are captured from a real city market. The person images are cropped from original images by the DPM detector~\cite{felzenszwalb2010object}. 2) CUHK03~\cite{li2014deepreid} contains 14,097 images of 1,467 persons captured by two cameras on a campus. The person images are manually cropped from the scene images. 3) DukeMTMC-ReID~\cite{ristani2016MTMC} is also collected from a campus. Manually drawn bounding boxes are used to crop person images from the surveillance images. we follow the setup in \cite{zheng2017unlabeled} to divide DukeMTMC-ReID dataset into train and test splits, which contain 16,522 images of 702 persons for training and 18,363 images of other 702 persons for testing.

\textbf{Evaluation metrics.} Mean average precision (mAP) and CMC top-1, top-5, top-10 accuracies are adopted as evaluation metrics. For each dataset, different mAP, and CMC computation methods are used following their original setup to calculate the final performance.

%
%

\begin{table*}[tb]
   \small
   \setlength{\tabcolsep}{5pt}
   \begin{center}
      \begin{tabular}{ccccccccccccccc}
         \toprule
         \multicolumn{3}{c}{Components}&
         \multicolumn{4}{c}{Market-1501~\cite{zheng2015scalable}} &
         \multicolumn{4}{c}{CUHK03~\cite{li2014deepreid}} &
         \multicolumn{4}{c}{DukeMTMC~\cite{ristani2016MTMC}}\\
         \#groups& RW &shuffle&mAP&top-1&top-5&top-10&mAP&top-1&top-5&top-10&mAP&top-1&top-5&top-10\\
         \midrule
         1 &$\times$ &$\times$ &76.4&\textbf{91.2}&\textbf{97.1}&\textbf{98.2}&88.9&91.1&97.6&98.7&61.8&78.8&88.5&91.0\\
         2 &$\times$ &$\times$&77.5&91.1&97.1&98.2& 90.0&92.2&98.2&98.9&62.6&\textbf{79.2}&88.7&91.0\\
         4 &$\times$ &$\times$ &\textbf{77.7}&91.1&96.9&97.9&\textbf{91.6}&\textbf{93.0}&\textbf{98.8}&\textbf{99.3}&\textbf{62.7}&78.9&\textbf{88.7}&\textbf{91.2}\\
         \hline

         1 & $\surd$ & $\times$ & 81.4&91.4&96.8&98.2& 91.5&92.4&97.4&98.8& 65.2&79.2&88.8&91.1 \\
         2 &$\surd$ &$\times$ &81.4&91.5&97.0&98.0&91.4&92.3&97.0&98.5&65.2&79.0&88.4&91.1\\
         4 &$\surd$ &$\times$ &\textbf{81.6}&\textbf{91.5}&\textbf{97.2}&\textbf{98.3}&\textbf{92.9}&\textbf{93.8}&97.3&98.2&\textbf{65.4}&\textbf{79.7}&\textbf{88.9}&\textbf{91.4}\\
         \hline
         2 &$\surd$ &$\surd$ &82.0&91.8&96.9&98.0& 93.1&93.9&98.2&99.0&65.4&78.6&88.1&90.8\\
         4 &$\surd$ &$\surd$ & \textbf{82.5}&\textbf{92.7}&96.9&98.1&\textbf{94.0}&\textbf{94.9}&\textbf{98.7}&\textbf{99.3}&\textbf{66.4}&\textbf{80.7}&88.5&90.8\\

         \bottomrule
      \end{tabular}
   \end{center}
   \vspace{-5pt}
   \caption{Ablation studies on the Market-1501~\cite{zheng2015scalable}, CUHK03~\cite{li2014deepreid} and DukeMTMC~\cite{ristani2016MTMC} datasets with different numbers of feature groups, end-to-end random walk (RW), and group-shuffle.}
   \label{tab:ablation}
   \vspace{-15pt}
\end{table*}

\begin{table}
   \small
   \begin{center}
      \begin{tabular}{p{3.5cm}<{\raggedright}p{0.7cm}<{\centering}p{0.7cm}<{\centering}p{0.7cm}<{\centering}p{0.7cm}<{\centering}}
         \toprule
         \multirow{2}{*}{Model}&
         \multicolumn{2}{c}{Market-1501}&
         \multicolumn{2}{c}{CUHK03}\\
          ~&mAP &top-1&mAP&top-1\\
         \midrule
         baseline &\textbf{76.4}&\textbf{91.2}&\textbf{88.9}&\textbf{91.1} \\
         baseline+triplet \cite{hermans2017defense} &68.3 &84.5 &-&- \\
         \hline
         baseline+dropout & 77.6&91.3&89.1&91.2\\
         baseline+group &77.7&91.1&91.6&93.0\\
         baseline+group+dropout &\textbf{78.1}&\textbf{91.3}&\textbf{91.3}&\textbf{93.3}\\
         \hline
         baseline+k-reciprocal \cite{Zhong_2017_CVPR} &78.5 &91.5 &89.9 & 92.2 \\
         baseline+RW w/o train &79.2 &\textbf{91.5} &90.2 &92.3 \\
         baseline+random walk & \textbf{81.4} & 91.4 & \textbf{91.5} & \textbf{92.4} \\
         \bottomrule
      \end{tabular}
   \end{center}
   \vspace{-5pt}
   \caption{Results of using the improved triplet loss \cite{hermans2017defense}, dropout \cite{srivastava2014dropout} and proposed feature grouping on the Market-1501~\cite{zheng2015scalable} and CUHK03~\cite{li2014deepreid} datasets}
   \label{tab:dropout}
   \vspace{-10pt}
\end{table}

\subsection{Implementation details} 
\label{sub:implementation_details}

The pairwise affinity CNN in our network adopts the ResNet-50 \cite{he2016deep} network structure and is pretrained on the ImageNet dataset. 
All the input person images are resized to $256 \times 128$. For data augmentation, random horizontal flipping and random erasing~\cite{zhong2017random} are adopted. We empirically set group number $K=4$ and $\lambda=0.95$ for our final model. The network is trained in an end-to-end manner with Stochastic Gradient Descent (SGD). For each mini-batch, we randomly sample training images according to their person identities. There are 64 persons' images in each batch and each person has 4 images, resulting in a batch size of 256. Among the images of each identity, we randomly choose 1 image as the probe and the remaining 3 images are used as gallery images. Note that the 192 gallery images are shared by all probe images. The initial learning rate is set to $10^{-4}$ and decreases to $10^{-5}$ after 50 epochs. The training generally converges after another 50 epochs. In testing, given a probe image, we first utilize the trained pairwise affinity CNN to identify the top-75 gallery images. The group-shuffling random walk operation is then utilized to refine the P2G affinities with their G2G affinities. The $K=4$ groups of refined P2G affinities are averaged as the final results. When random walk is not allowed in testing (\eg, for evaluating the learned person features), we directly use P2G affinities estimated by the pairwise affinity CNN for person image ranking. 

\begin{table}
   \small
   \setlength{\tabcolsep}{5pt}
   \begin{center}
      \begin{tabular}{ccccccc}
         \toprule
         \multicolumn{3}{c}{Components}&
         \multicolumn{4}{c}{Market-1501~\cite{zheng2015scalable}} \\
         \#groups& RW &shuffle&mAP&top-1&top-5&top-10\\
         \midrule
         1 &$\times$ &$\times$ &74.6&90.4&\textbf{96.9}&98.1\\
         2 &$\times$ &$\times$ &74.7&90.0&\textbf{96.6}&98.1\\
         4 &$\times$ &$\times$ &\textbf{75.9}&\textbf{90.5}&\textbf{96.9}&\textbf{98.3}\\
         \hline
         1 & $\surd$ & $\times$&75.6&90.8&97.0&\textbf{98.2}\\
         2 &$\surd$ &$\times$& 76.3&91.2&\textbf{97.1}&\textbf{98.2}\\
         4 &$\surd$ &$\times$ &\textbf{76.7}&\textbf{91.4}&96.9&\textbf{98.2}\\
         \hline
         2 &$\surd$ &$\surd$ &\textbf{77.0}&\textbf{91.3}&97.1&98.2\\
         4 &$\surd$ &$\surd$ &76.9&\textbf{91.3}&\textbf{97.3}&\textbf{98.4}\\

         \bottomrule
      \end{tabular}
   \end{center}
   \vspace{-5pt}
   \caption{Results of estimating P2G affinities as feature distances by our trained ResNet-50 on the Market-1501~\cite{zheng2015scalable} dataset.}
   \label{tab:ablation_feat_market}
   \vspace{-10pt}
\end{table}

\begin{table}
   \small
   \setlength{\tabcolsep}{5pt}
   \begin{center}
      \begin{tabular}{ccccccc}
         \toprule
         \multicolumn{3}{c}{Components}&
         \multicolumn{4}{c}{DukeMTMC~\cite{ristani2016MTMC}}\\
         \#groups& RW &shuffle&mAP&top-1&top-5&top-10\\
         \midrule
         1 &$\times$ &$\times$ &60.3&77.6&\textbf{87.6}&90.1\\
         2 &$\times$ &$\times$& 60.4&77.2&87.3&90.2\\
         4 &$\times$ &$\times$ &\textbf{61.5}&\textbf{77.7}&87.5&\textbf{90.4}\\
         \hline
         1 & $\surd$ & $\times$&60.8&77.8&87.6&90.4\\
         2 &$\surd$ &$\times$&61.0&77.8&\textbf{87.7}&90.3\\
         4 &$\surd$ &$\times$ &\textbf{61.7}&\textbf{77.9}&87.6&\textbf{90.5}\\
         \hline
         2 &$\surd$ &$\surd$&61.2&77.7&\textbf{88.0}&\textbf{90.8}\\
         4 &$\surd$ &$\surd$ &\textbf{62.1}&\textbf{78.1}&87.8&90.3\\

         \bottomrule
      \end{tabular}
   \end{center}
   \vspace{-5pt}
   \caption{Results of estimating P2G affinities as feature distances by our trained ResNet-50 on the DukeMTMC~\cite{ristani2016MTMC} dataset.}
   \label{tab:ablation_feat_duke}
   \vspace{-15pt}
\end{table}

\subsection{Ablation study} 
\label{sub:ablation_study}

In this section, we investigate the effectiveness of each component in our proposed group-shuffling random walk by conducting a series of experiments on the Market-1501, CUHK03, and DukeMTMC datasets. 

\textbf{Baseline model and comparison with triplet loss \cite{hermans2017defense}.} We utilize the pairwise affinity CNN in our framework with the group number $K=1$ as our baseline model. To fully utilize all available information in each mini-batch of size $256$, unlike our final model that uses only 256 ground-truth P2G affinity scores as training supervisions,  $64 \times 192$ P2G pairs and all $192^2$ G2G pairs are used for training. The P2G-to-G2G ratio is therefore 1:3. We compare the baseline with verification loss to the same ResNet-50 structure with the improved triplet loss \cite{hermans2017defense} (denoted by baseline+triplet). Results in Table \ref{tab:dropout} show our baseline outperforms the state-of-the-art triplet loss by 13.7\% in terms of mAP with the same ResNet-50 structure.

\textbf{Feature grouping versus/plus dropout.} We investigate the influence of feature grouping on Market-1501 and CUHK03 datasets, and compare/combine it with the feature dropout techqniue \cite{srivastava2014dropout} (see Table \ref{tab:dropout}). We first test only applying dropout to the features of the last FC layer in the baseline (denoted by baseline+dropout). Note that we set different dropout ratios to the two datasets, \ie, 0.5 for Market-1501 and 0.3 for CUHK03, to achieve the optimal performance. As shown by the results, the dropout leads to marginal improvements. We then test dividing the 2048-d feature vector into $K=4$ feature groups and applying the P2G affinity supervisions to each of them (denoted by baseline+group). It results in better performance than baseline+dropout. We further combine the proposed feature grouping with dropout (denoted by baseline+group+dropout). The results show further improvements over baseline+group and demonstrate that the feature grouping is complementary with feature dropout.

\textbf{Comparison with re-ranking as post-processing.} We then compare our approach with k-reciprocal re-ranking \cite{Zhong_2017_CVPR} that treats the re-ranking as a separate post-processing step. For fair comparison, we implement k-reciprocal re-ranking with the features learned by our baseline model. The initial affinities are calculated as pairwise feature distances. The performance outperforms the original results in the paper but is inferior to our proposed approach with end-to-end random walk without feature grouping and group-shuffling (denoted by baseline+RW). To validate the effectiveness of end-to-end random walk for training, we apply a separate random walk operation as post-processing to the affinity scores from our baseline (denoted by baseline+RW w/o train). The performance outperforms k-reciprocal re-ranking but is still inferior to our end-to-end approach, which demonstrates the importance of training the deep neural network with end-to-end random walk operation.

\textbf{Feature group number $K$.} We then investigate the influence of different feature group numbers $K$. As shown by in Table \ref{tab:ablation}, utilizing 4 feature groups generally outperforms those without feature groups by $\sim$1\% in terms of mAP.

\textbf{Random walk with feature grouping.} When incorporating the random walk operation with no group-shuffling into the network (rows 4-6 in Table \ref{tab:ablation}), the mAP on Market-1501, CUHK03, DukeMTMC datasets increase by 2.9\%, 1.3\%, and 2.7\%. Grouping to 4 feature sub-vectors improve the mAP on CUHK03 by 1.4\% but shows marginal improvements on Market-1501 and DukeMTMC datasets.


\textbf{Group-shuffling random walk.} For validating the effectiveness of group-shuffling, we conduct random walk with group-shuffling as described in Section \ref{ssec:group_shuffling_random_walk}. Note that $K=2$ and $4$ results in 4 and 16 groups of refined P2G affinities for applying supervisions. Comparing results in rows 5-6 and rows 7-8 of Table \ref{tab:ablation} shows the group-shuffling operation with $K=4$ generally brings $\sim$1\% improvements in terms of mAP on the three datasets.

\textbf{Better features with group-shuffling random walk.} Our approach does not only improve the final accuracy on person re-ID, but also learns better person features via the proposed group-shuffling random walk.  To show this, we directly utilize the trained ResNet-50 from our network to extract visual features of the test images. Image pairwise affinities are estimated as the Euclidean distances between them. The results on Market-1501 and DukeMTMC datasets are recorded in Tables \ref{tab:ablation_feat_market}-\ref{tab:ablation_feat_duke}, which show that all our proposed operations, \ie, feature grouping, end-to-end random walk, and group shuffling, contribute to learning better visual features. Incorporating the proposed operations in the testing phase could further boost the final accuracy (Tables \ref{tab:ablation} vs. \ref{tab:ablation_feat_market}-\ref{tab:ablation_feat_duke}).

\begin{table}
   \small
   \setlength{\tabcolsep}{3pt}
      \begin{center}
      \begin{tabular}{llcccc}
         \toprule
         \multirow{2}{*}{Methods}&
         \multirow{2}{*}{Reference}&
         \multicolumn{4}{c}{Market-1501~\cite{zheng2015scalable}}\\
         & &mAP&top-1&top-5&top-10\\
         \midrule
         OIM Loss~\cite{xiao2017joint}&CVPR 2017& 60.9 & 82.1  &~ - &~ - \\
         CADL~\cite{Lin_2017_CVPR} & CVPR 2017&47.1&73.8&~ - &~ - \\
         P2S~\cite{Zhou_2017_CVPR} &CVPR 2017&44.3&70.7&~ - &~ - \\
         MSCAN~\cite{Li_2017_CVPR} &CVPR 2017&53.1&76.3&~ - &~ - \\
         SSM~\cite{bai2017scalable}&CVPR 2017&68.8&82.2&~ - &~ -\\
         DCA~\cite{li2017learning}&CVPR 2017&57.5&80.3&~ -&~ -\\
         SpindleNet~\cite{zhao2017spindle}&CVPR 2017&~ -&76.9&91.5&94.6\\
         k-reciprocal~\cite{Zhong_2017_CVPR}&CVPR 2017& 63.6&77.1&~ - &~ - \\
         VI+LSRO~\cite{zheng2017unlabeled} &ICCV 2017&66.1&84.0&~ -&~ -\\
         OL-MANS~\cite{Zhou_2017_ICCV} &ICCV 2017&~ -&60.7&~ -&~ -\\
         PDC~\cite{Su_2017_ICCV}&ICCV 2017&63.4&84.1&92.7&94.9\\
         PA~\cite{zhao2017deeply}&ICCV 2017& 63.4 & 81.0 &92.0&94.7\\
         SVDNet~\cite{Sun_2017_ICCV}&ICCV 2017& 62.1&82.3&92.3&95.2\\
         JLML~\cite{gong2017person}&IJCAI 2017&65.5&85.1&~ - &~ -\\
         Proposed &~& \textbf{82.5} &\textbf{92.7} &\textbf{96.9} & \textbf{98.1}\\
         \bottomrule
      \end{tabular}
   \end{center}
   \vspace{-5pt}
   \caption{mAP, top-1, top-5, and top-10 accuracies of compared methods on the Market-1501 dataset~\cite{zheng2015scalable}.
   }
   \label{tab:market}
   \vspace{-15pt}
\end{table}

\begin{table}
\setlength{\tabcolsep}{3pt}
   \small
   \begin{center}
      \begin{tabular}{llcccc}
         \toprule
         \multirow{2}{*}{Methods}&
         \multirow{2}{*}{Reference}&
         \multicolumn{4}{c}{CUHK03~\cite{li2014deepreid}}\\
         & &mAP&top-1&top-5&top-10\\
         \midrule
         OIM Loss~\cite{xiao2017joint} &CVPR 2017& 72.5 & 77.5  &~ - &~ - \\
         MSCAN~\cite{Li_2017_CVPR} &CVPR 2017&~ -&74.2&94.3&97.5\\ 
         DCA~\cite{li2017learning} &CVPR 2017&~ -&74.2&94.3&97.5\\
         SSM~\cite{bai2017scalable}&CVPR 2017&~ -&76.6&94.6 & 98.0\\
         SpindleNet~\cite{zhao2017spindle} &CVPR 2017&~ -&88.5&97.8 & 98.6\\
         k-reciprocal~\cite{Zhong_2017_CVPR}&CVPR 2017&67.6 &61.6&~ - &~ - \\
         Quadruplet~\cite{Chen_2017_CVPR}&CVPR 2017&~ -&75.5&95.2&99.2\\
         OL-MANS~\cite{Zhou_2017_ICCV}&ICCV 2017&~ -&61.7&88.4&95.2\\
         PA~\cite{zhao2017deeply}&ICCV 2017&~ -& 85.4 &97.6&99.4\\
         SVDNet~\cite{Sun_2017_ICCV} &ICCV 2017&84.8& 81.8&95.2&97.2\\
         VI+LSRO~\cite{zheng2017unlabeled}&ICCV 2017&87.4&84.6&97.6&98.9\\
         PDC~\cite{Su_2017_ICCV}&ICCV 2017&~ -&88.7&98.6&\textbf{99.6}\\
         MuDeep~\cite{qian2017multi}&ICCV 2017&~ -&76.3 &96.0 &98.4\\
         JLML~\cite{gong2017person}&IJCAI 2017&~ -&83.2&98.0 & 99.4\\
         Proposed & ~&\textbf{94.0} &\textbf{94.9} & 98.7 & 99.3\\
         \bottomrule
      \end{tabular}
   \end{center}
   \vspace{-5pt}
   \caption{mAP, top-1, top-5, and top-10 accuracies by compared methods on the CUHK03 dataset~\cite{li2014deepreid}.}
   \label{tab:cuhk}
   \vspace{-15pt}
\end{table}

\begin{table}{\hspace{-1.0cm}}
   \small
   \setlength{\tabcolsep}{3pt}
   \begin{center}
      \begin{tabular}{llcccc}
         \toprule
         \multirow{2}{*}{Methods}&
         \multirow{2}{*}{Reference}&
         \multicolumn{4}{c}{DukeMTMC~\cite{ristani2016MTMC}}\\
         & &mAP&top-1&top-5&top-10\\
         \midrule
         BoW+KISSME~\cite{zheng2015scalable} &ICCV 2015& 12.2 &25.1&~ -&~ -\\
         LOMO+XQDA~\cite{liao2015person} &CVPR 2015& 17.0 &30.8&~ -&~ -\\
         OIM Loss~\cite{xiao2017joint} &CVPR 2017& 47.4 & 68.1  &~ - &~ - \\
         ACRN~\cite{schumann2017person}&CVPR 2017& 52.0 &72.6&84.8&88.9\\
         OIM Loss~\cite{xiao2017joint}&CVPR 2017& 47.4 & 68.1  &~ - &~ - \\
         Basel+LSRO~\cite{zheng2017unlabeled} &ICCV 2017&47.1&67.7&~ -&~ -\\
         SVDNet~\cite{Sun_2017_ICCV}&ICCV 2017&56.8 &76.7&86.4&89.9\\
         
         Proposed  & ~&\textbf{66.4} &\textbf{80.7} &\textbf{88.5} & \textbf{90.8}\\
         \bottomrule
      \end{tabular}
   \end{center}
   \vspace{-5pt}
   \caption{mAP, top-1, top-5, and top-10 accuracies by compared methods on the DukeMTMC dataset~\cite{ristani2016MTMC}.}
   \label{tab:duke}
   \vspace{-15pt}
\end{table}

\subsection{Comparison with state-of-the-art methods} 
\label{sub:compared_with_state_of_art_methods}

\textbf{Results on Market-1501 dataset.} Table \ref{tab:market} shows the results of our proposed group-shuffling random walk approach and state-of-the-art methods on the Market-1501 dataset. Our approach outperforms all compared methods in terms of meanAP, top-1, top-5,  and top-10 accuracies, which demonstrate the effectiveness of the proposed method on this dataset.

Consistent-aware deep learning~\cite{Lin_2017_CVPR} (CADL) aims to obtain the maximal correct matches for the whole camera network. 
It regularizes the matching results of a probe image to be similar across different cameras. Compared with CADL, our approach improves by 35.4\% and 18.9\% in terms of meanAP and top-1 accuracy.
Supervised Smoothed Manifold (SSM) ~\cite{bai2017scalable} utilized random walk operation as a post-processing stage during testing, which estimates the similarity value between two instances in the context of other pairs of instances. Our approach outperforms SSM by 13.7\% and 20.5\% in terms of meanAP and top-1 accuracy.
k-reciprocal encoding rerank (k-reciprocal)~\cite{Zhong_2017_CVPR} encoded each probe image's k-reciprocal nearest neighbors into a single vector, which is utilized for re-ranking under the Jaccard distance. Our approach outperforms k-reciprocal by 18.9\% and 15.4\% in terms of meanAP and top-1 accuracy. 
Unlike existing methods that learn a single global metric for all probes. Online local metric adaptation (OL-MANS) exploits negative samples to learn a dedicated local metric for each online probe. Our proposed method outperforms OL-MANS by 32.0\% in terms of top-1 accuracy. 

\textbf{Results on CUHK03 dataset.}
The Re-ID results on CUHK03 dataset is shown in Table \ref{tab:cuhk}.   The meanAP and top-1 accuracy of our framework are 94.0\% and 94.9\%, which outperform those by state-of-the-art methods. For top-10 accuracy, PDC~\cite{Su_2017_ICCV} yields slightly better performance than ours. However, PDC needs human pose information for better aligning visual features, which is not utilized in our framework. SpindleNet~\cite{zhao2017spindle} and PA~\cite{zhao2017deeply} also utilize similar human pose information.
The gain on top-1 accuracy by our method is 19.4\% compared to Quadruplet loss~\cite{Chen_2017_CVPR}, which aims to enforce the minimum inter-class distance being greater than the maximum intra-class distance in  sampled quadruplets. 
MuDeep~\cite{qian2017multi} utilized a GoogLeNet-like~\cite{szegedy2015going} structure to learn discriminative features with different spatial scales and locations of person images. Our method improves the top-1 accuracy by 18.6\% compared with MuDeep.
Verif-Identif.+LSRO (VI+LSRO)~\cite{zheng2017unlabeled} utilizes additional training data generated by GAN, while our method does not utilize any additional training data but still outperforms it.

\textbf{Results on DukeMTMC dataset.}
In Table \ref{tab:duke}, we show the results of our framework and those by state-of-the-art ones on the DukeMTMC dataset. Our method outperforms all the compared frameworks. Compared with the state-of-the-art SVDNet~\cite{Sun_2017_ICCV}. The gains on meanAP and top-1 accuracy by our proposed framework are 9.6\% and 14.0\% respectively.


\section{Conclusion}
\label{sec:conclusion}
In this paper, we proposed a novel group-shuffling random walk operation for fully utilizing the affinities between gallery images (G2G affinities) to refine the affinities between probe and gallery images (P2G affinities). Compared with the previous re-ranking methods, our approach integrates random walk operation into the training process of deep neural networks. Furthermore, by grouping and shuffling the features, discriminative person features could be learned with rich supervisions. The overall performance of our approach outperforms baseline methods and state-of-the-art approaches by large margins, which demonstrates the effectiveness of our proposed approach.


{\small
\bibliographystyle{ieee}
\bibliography{egbib}
}

\end{document}